\documentclass[sigconf]{acmart}
\usepackage{tabularx} 
\usepackage[utf8]{inputenc}
\usepackage[T1]{fontenc}
\usepackage{balance}

\AtBeginDocument{%
  \providecommand\BibTeX{{%
    \normalfont B\kern-0.5em{\scshape i\kern-0.25em b}\kern-0.8em\TeX}}}

\copyrightyear{2024} 
\acmYear{2024} 

\setcopyright{acmlicensed}
\acmConference[WWW '24 Companion]{Companion
Proceedings of the ACM Web Conference 2024}{May 13--17, 2024}{Singapore,
Singapore}
\acmBooktitle{Companion Proceedings of the ACM Web Conference 2024 (WWW
'24 Companion), May 13--17, 2024, Singapore, Singapore}
\acmDOI{10.1145/3589335.3651584}
\acmISBN{979-8-4007-0172-6/24/05}

\begin{document}

\title{Zero-shot Explainable Mental Health Analysis on Social Media by Incorporating Mental Scales}
\author{Wenyu Li}

\authornote{Corresponding author}
\email{202130131566@mail.scut.edu.cn}
\orcid{0009-0009-2792-4303}

\affiliation{%
  \institution{South China University of Technology}
  \city{Guangzhou}
  \state{Guangdong}
  \country{China}
}

\author{Yinuo Zhu}
\email{202164690589@mail.scut.edu.cn}

\affiliation{%
  \institution{South China University of Technology}
  \city{Guangzhou}
  \state{Guangdong}
  \country{China}
}
\author{Xin Lin}
\email{202164690367@mail.scut.edu.cn}

\affiliation{%
  \institution{South China University of Technology}
  \city{Guangzhou}
  \state{Guangdong}
  \country{China}
}
\author{Ming Li}
\email{202164690343@mail.scut.edu.cn}

\affiliation{%
  \institution{South China University of Technology}
  \city{Guangzhou}
  \state{Guangdong}
  \country{China}
}
\author{Ziyue Jiang}
\email{202130200293@mail.scut.edu.cn}

\affiliation{%
  \institution{South China University of Technology}
  \city{Guangzhou}
  \state{Guangdong}
  \country{China}
}
\author{Ziqian Zeng}
\authornotemark[1]
\email{zqzeng@scut.edu.cn}

\affiliation{%
  \institution{South China University of Technology}
  \city{Guangzhou}
  \state{Guangdong}
  \country{China}
}

\renewcommand{\shortauthors}{Wenyu Li, et al.}

\begin{abstract}
Traditional discriminative approaches in mental health analysis are known for their strong capacity but lack interpretability and demand large-scale annotated data. The generative approaches, such as those based on large language models (LLMs), have the potential to get rid of heavy annotations and provide explanations but their capabilities still fall short compared to discriminative approaches, and their explanations may be unreliable due to the fact that the generation of explanation is a black-box process. 
Inspired by the psychological assessment practice of using scales to evaluate mental states, our method which is called \textbf{M}ental \textbf{A}nalysis by \textbf{I}ncorporating \textbf{M}ental \textbf{S}cales (MAIMS), incorporates two procedures via LLMs. First, the patient completes mental scales, and second, the psychologist interprets the collected information from the mental scales and makes informed decisions. 
Experimental results show that MAIMS outperforms other zero-shot methods. 
MAIMS can generate more rigorous explanation based on the outputs of mental scales. 
\footnote{More detail information see: \url{https://github.com/w-y-li/MAIMS.git}}

\end{abstract}




\maketitle
\begin{figure*}[t]
  \centering
  \includegraphics[width=0.6\textwidth]{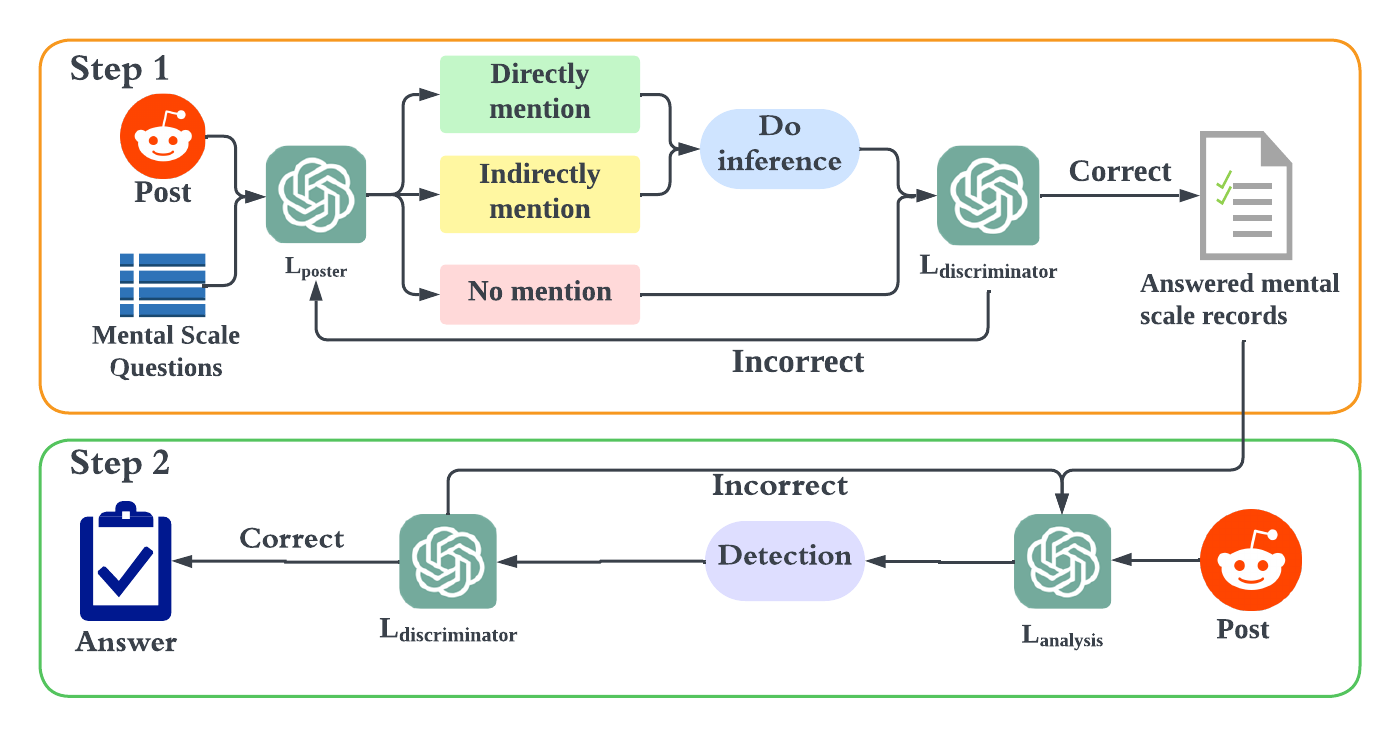}
  \caption{Two-step workflow of MAIMS. Step 1 uses a language model to interpret posts against mental scales, verified by a discriminator. In Step 2, a language model assesses mental health, with a second discriminator confirming the accuracy.}
  \label{fig:flowchart}
\end{figure*}

\section{Introduction}
Mental health issues are a growing public health concern globally \cite{r42}. The past decade has witnessed a significant surge in efforts to evaluate symptomatology linked with conditions like depressive disorders, self-harm, and the severity of mental illness, utilizing non-clinical data sources \cite{r28}. With the increasing accessibility of the internet and the rising usage of social media platforms like Reddit and Twitter, the content expressing emotion has been widely utilized as a resource for analyzing mental health conditions. This trend potentially facilitates the development of methodologies for detecting mental health problems among social media users. Advancement in Large Language Models (LLMs) and datasets, have led to sophisticated models like MentaLLaMA \cite{r10}, marking a significant enhancement in the field. 

However, existing approaches to mental health analysis on social media have encountered several challenges, such as intensive annotation for training data, limited interpretability and lack of scientific basis. 
While some LLMs can generate explanations \cite{r10}, the foundation for their interpretations remains unclear.
To bridge these gaps, we have drawn inspiration from the real-life psychological counseling process, whereby individuals are often initially asked to complete self-rating mental scales. We replicate this procedure through two steps: Firstly, we employ an LLM to mimic the user completing mental scales. Secondly, we utilize another LLM to leverage the information obtained from these scales for an improved analysis of users' mental well-being. Our approach aims to provide a transparent and reliable means of mental health analysis through integrating the scientifically grounded nature of mental scales as enhancement of our detection process while exploring the latent ability of LLMs to infer individual personality traits (e.g. anxiety-related personality) from text. Our results demonstrate that the completed scales contribute to a better understanding and inference of user posts, ultimately improving the accuracy and interpretability of final decision.

Our main contributions are summarized as follows: 
\begin{itemize}
    \item  We introduce a novel methodology for detecting mental health issues, with the first attempt to answer mental scales automatically based on user-text content. 
    \item  We explore the potential of LLMs to infer individual personality traits from text, which reveals the capability of LLMs to simulate human-like completion of mental scales with minimal user data. That enhances the scientific nature of mental health detection without relying on training labels. 
    \item Our approach show a comparative performance method of all method, outperforms other zero-shot and fine-tuning methods, with nearly 16\% higher weighted F1-score on IRF task compare to baseline zero shot method.
\end{itemize}

\section{Method}
MAIMS for mental health analysis is framed as Mental Scale Enhanced Mental Health Classification Tasks, as illustrated in Fig. \ref{fig:flowchart}. There are two steps to address the mental health classification task: Mental scales completion and mental health analysis. This approach leverages three foundational auto-regressive language model $L_{poster}$, $L_{analysis}$ and $L_{discriminator}$, where $L_{poster}$ acts as the poster and $L_{analysis}$ analyzes mental health issues. To mitigate challenges encountered by the language model, such as negative inference (i.e. a cognitive bias where an individual tends to interpret content in a negative light, even when the actual content may not be explicitly negative.) and misunderstanding of the content (e.g. the language model may mistakenly think the content is describing the poster himself while it is actually describing the poster's father), we incorporates $L_{discriminator}$ as an as a discriminator to evaluate the quality of mental scale completion and the final mental health analysis. 

\textbf{Step 1: Mental scales completion. } $L_{poster}$ processes a set of inputs $\{(p_i, s_{undo})\}_{i=1,\cdots,N}$ to imitate the poster completing the mental scale, where $p$ denotes a social media post, $s$ represents a sequence of questions from a specific mental scale and the corresponding criteria, $N$ is the total number of posts. To begin, $L_{poster}$ classifies posts into three categories: direct mention, indirect mention and no mention of the questions in the scales. Subsequently, if scale-related issues are identified, further inference is carried out to complete the mental scales, with answers captured in $s_{done}$. Then the discriminator $L_{discriminator}$ evaluates the quality of the completed scales based on the input set $\{(p_i, s_{undo}, s_{done})\}_{i=1,\cdots,N}$. Only when the answers are considered correct by $L_{discriminator}$ will be saved in $s_{done}$, otherwise, the scale completing process will be repeated.

\textbf{Step 2: Mental health analysis. } $L_{analysis}$ processes a set of inputs $\{(s_{done}, p_{i})\}_{i=1,...N}$ and detects mental health issues in posts. $r_{i}$ is the detection result and it will be forwarded to the discriminator $L_{discriminator}$ alongside the original inputs: $\{(r_{i},s_{done}, p_{i})\}_{i=1,...N}$. $L_{discriminator}$evaluates the detection's quality by analyzing whether $L_{analysis}$ has made any over-inferences or other mistakes. If the results are deemed accurate and acceptable, $r_{i}$ is utilized as the final output $M_d$.

For enhancing interpretablity, we prompt the model to give a reason concurrently to help us understand the whole process and show its scientific induction while generating all the outputs above.

\section{Experiment and Analysis}
\subsection{Datasets and Mental Scales Description}
We conduct experiments using two datasets, Depression\_Reddit (DR) \cite{r30} and IRF, and utilize the Beck Depression Inventory (BDI) and the Interpersonal Needs Questionnaire (INQ-15) as mental scales respectively. The DR dataset and BDI aim to detect symptoms associated with a specific mental health condition, while the IRF dataset and INQ-15 focuse on identifying the underlying social and psychological factors related to mental health.

\subsection{Baseline Models}
We have meticulously curated a range of robust baseline models to serve as benchmarks for comparison with our own models. These include discriminative methods, such as fine-tuning pre-trained language models like RoBERTa \cite{r5} and BERT \cite{r4}, as well as SOTA models, Mental-RoBERTa and MentalBERT \cite{r6}.

For interpretable mental health analysis methods, we have employed zero-shot methods utilizing LLaMA-7B, LLaMA-13B \cite{r35}, ChatGPT and GPT-3.5. Additionally, we have adapted BART-large \cite{r37}, T5-large \cite{r38} and LLaMA-7B to assess fine-tuning methods. Moreover, we have integrated instruction-tuning methods into our exploration by leveraging the MentaLLaMA \cite{r10} series, including MentaLLaMA-7B as well as the 7B and 13B iterations of MentaLLaMA-chat.

\subsection{Results Analysis}
\textbf{Main Results.}
The evaluation results in Table \ref{Table: Main results}, measured in the Weighted F1-score, reveal several key insights: 1) MAIMS achieves SOTA performance in zere-shot methods, underscoring the advantages of incorporating Mental Scale augmentation. 2) MAIMS outperforms supervised training methods using generative models without requiring training labels or fine-tuning. 3) MAIMS outperforms the discriminative methods on the IRF dataset, but there is still a gap in performance on the DR Dataset. 

\noindent \textbf{Interpretability.} Here, we take a closer look into our interpretable and scientific process of analysis. Specifcally, given a certain user post: \textit{``The good times don't make the bad times worth living... Even the good times are tainted and I can barely ever even enjoy them the way I'm supposed to...''}, MAIMS works effectively by generating a response anchored to the Mental Scale: \textit{``In the post, I mentioned that even the good times are tainted and I can barely ever enjoy them. This implies that I often feel sad.''}, which reasonably infers \textit{``I feel sad much of the time''} in response to question 1 in the BDI Scale, leading to the accurate identification of the sadness symptom. Notably, the response refers to the original text from the initial post, ensuring a scientific and traceable result throughout the process. By systematically inferring related symptoms from the scale questions, MAIMS enables reasonable detection and provides scientific explanations: ``\textit{Answer: Yes. The poster's responses to scale indicate symptoms of depression such as sadness, ..., Although the poster does not report any issues with sleep or appetite. The presence of multiple depressive symptoms suggests a potential diagnosis of depression.}''

\noindent \textbf{Personality traits inference.} As can be seen in Table \ref{Table: Example results}, the poster clearly mentions content such as \textit{``feel like not allowed to feel bad''} and \textit{``I'm an unimportant piece of ship''}. In the scale analysis of the model, the model reasonably infers the poster's own feelings mentioned in the post, analyzes the poster's psychology, and matches the poster's psychological state with scale questions 13 and 20.

\begin{table}[]

\caption{
Weighted F1 scores on the DR and IRF test set.
In the final block, subscript ``ZS'' denotes that LLMs perform mental health analysis in a zero-shot manner. 
}
\centering
\small
\label{Table: Main results}
\begin{tabular}{ccc}
\toprule
\multicolumn{1}{c|}{\textbf{Model}}               &\textbf{DR}                                  & \textbf{IRF}   \\ \bottomrule
                                         &\multicolumn{2}{l}{\textbf{Discriminative methods} }  \\
\multicolumn{1}{l|}{BERT-base\cite{r4}}           & 90.90                             & 72.30 \\
\multicolumn{1}{l|}{RoBERTa-base\cite{r5}}        & 95.11                               & 71.35 \\
\multicolumn{1}{l|}{MentalBERT\cite{r6}}          & 94.62                               & 76.73 \\
\multicolumn{1}{l|}{MentalRoBERTa\cite{r6}}       & 94.23                              & -     \\ \bottomrule
                                         &\multicolumn{2}{l}{\footnotesize\textbf{Completion-based fine-tuning methods}}    \\
\multicolumn{1}{l|}{T5-Large\cite{r38}}            & 84.90                                & 74.00  \\
\multicolumn{1}{l|}{BART-Large\cite{r37}}          & 84.60                                & 76.20  \\
\multicolumn{1}{l|}{LLaMA2-7B\cite{r35}}           & 84.94                               & 73.50  \\ \bottomrule
                                         &\multicolumn{2}{l}{\textbf{Instruction-tuning methods}}    \\
\multicolumn{1}{l|}{MentaLLaMA-7B\cite{r10}}       & 76.14                               & 67.53 \\
\multicolumn{1}{l|}{MentaLLaMA-chat-7B\cite{r10}}  & 83.95                              & 72.88 \\
\multicolumn{1}{l|}{MentaLLaMA-chat-13B\cite{r10}} & 85.68                               & 76.49 \\ \bottomrule
                                         &\multicolumn{2}{l}{\textbf{Zero-shot methods }}    \\
\multicolumn{1}{l|}{LLaMA-7B\textsubscript{ZS}\cite{r35}}          & 58.91                               & 38.02 \\
\multicolumn{1}{l|}{LLaMA-13B\textsubscript{ZS}\cite{r35}}          & 54.07                               &38.89 \\
\multicolumn{1}{l|}{ChatGPT\textsubscript{ZS} }          & 82.41                               & 41.33 \\
\multicolumn{1}{l|}{GPT3.5\textsubscript{ZS}}          & 86.66                               & 67.23 \\
\multicolumn{1}{l|}{\textbf{MAIMS}}          & \textbf{88.68}                               & \textbf{77.93} \\
\bottomrule
\end{tabular}
\end{table}

\begin{table}[]
\caption{Evaluation results on the initial 100 data points of DR and IRF test set. ``Without MS'' denotes without using mental scales.
``Without Discrm.'' denotes without using a Discriminator.}
\centering
\label{Table: Abalation results}
\begin{tabular}{lllllll}
\toprule
            &\multicolumn{2}{c}{\small{Without Discrim.}} 
            &\multicolumn{2}{c}{\small{Without MS}}   &\multicolumn{2}{c}{\small{\textbf{MAIMS}}}  \\ \hline
            & DR         & IRF & DR     & IRF & DR        & IRF \\ \hline
Weighted-F1 & 73.29          & 62.83   & 77.74      & 78.17   & \textbf{88.16}   &\textbf{81.63}   \\
Accuracy    & 79          & 72   & 76      & 79   & \textbf{88}         &\textbf{ 82}    \\ \bottomrule
\end{tabular}
\end{table}

\begin{table}[]
\centering
\caption{An example to shown the potential of LLMs to infer individual personality traits.}
\label{Table: Example results}
\begin{tabular}{lp{6.5cm}}
\toprule
\normalfont
Question:   & Does the poster suffer from depression?                                                                                                                       \\ \hline
True Label: & Yes                                                                                                                                                                \\
Post:       & ...... I feel like I'm not allowed to feel bad. I'm an unimportant piece of shit and there are so many other people that need help and all the other people are a lot more important than me. ...... It's like I'm stuck in a swamp and all the people step on me and push me deeper down just to get out themselves. ......
     \\
MS Record:  &  ('13', ['directly\_mention', 'I find it more difficult to make decisions than usual. The post mentions \textbf{feeling stuck} and \textbf{being pushed down}, which suggests a \textbf{lack of agency and difficulty in making decisions}."]) ('20', ['directly\_mention', 'I am too tired or fatigued to do most of the things I used to do.In the post, the poster states that \textbf{they feel like they are not allowed to feel bad}, and they \textbf{describe themselves as an unimportant piece of shit}. They also \textbf{mention feeling stuck in a swamp and being stepped on by others}. .......'])
\\
\bottomrule
\end{tabular}
\end{table}

           

\noindent \textbf{Ablation Study 1: Impact of  Mental Scales \footnote{Use the first 100 samples in this study due to the high cost of using Openai API.}.}
We prepare a variant of MAIMS by removing the mental scales completion task and report the performance comparison in Table \ref{Table: Abalation results}, which demonstrates a clear performance difference, substantiating the valuable role of mental scales in enhancing the model's judgment accuracy and efficiency. This enhancement is achieved by thoroughly extracting information from posts and establishing a scientific basis for judgment.

\begin{figure}[htbp]
  \centering
  \includegraphics[width=0.8\linewidth]{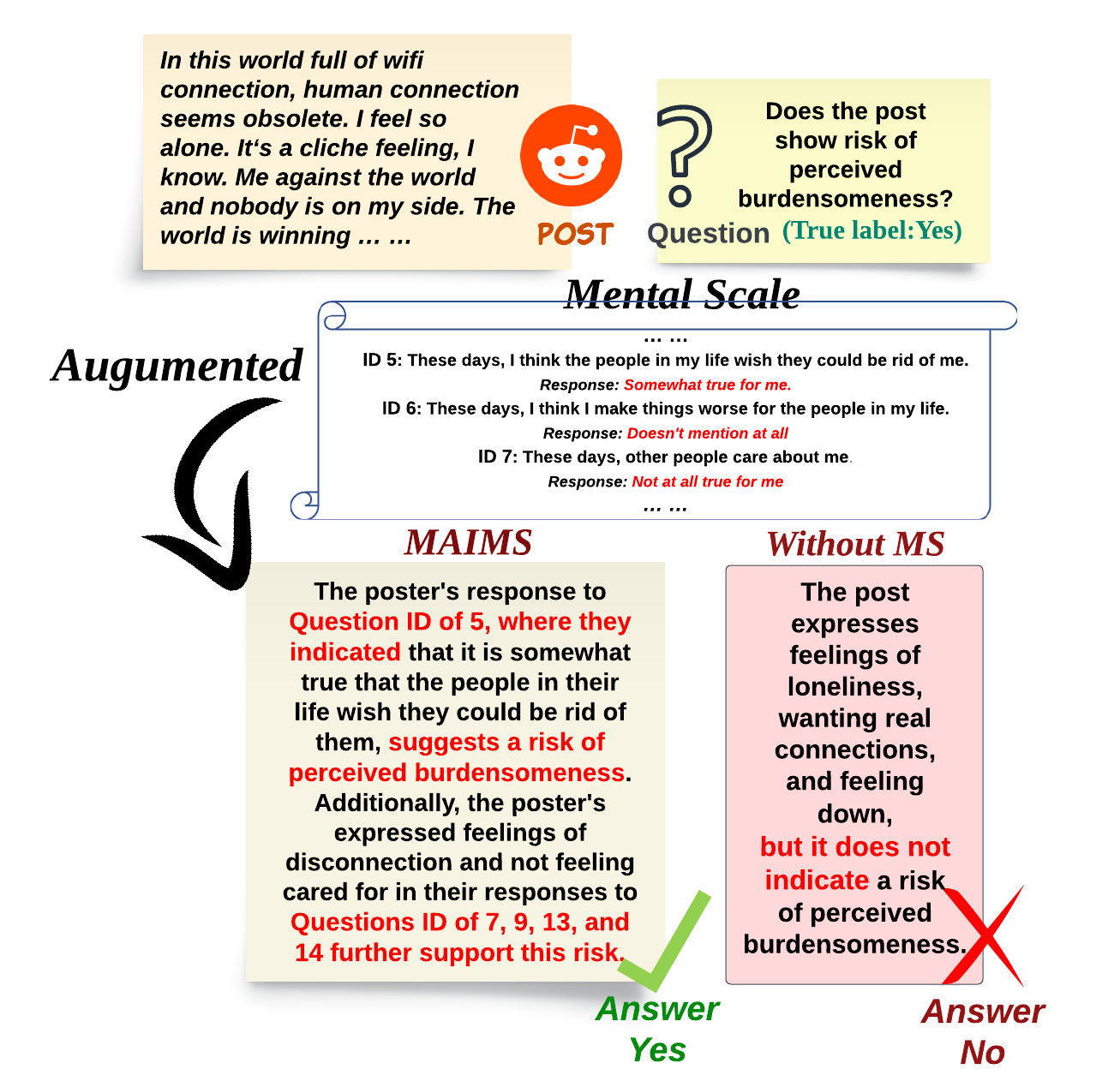}
  \caption{Ablation Study of Mental Scales: Example}
  \label{fig:Ablation MS}
\end{figure}

In our ablation study using the initial 100 data points from the IRF dataset, we chose post 59 to demonstrate the experimental process and outcomes, as depicted in Fig. \ref{fig:Ablation MS}. Without the aid of the mental scales, the model's judgment appears arbitrary, focusing solely on the general meaning of the post and failing to analyze the post's details comprehensively. Conversely, integration of the mental scales module allows the model to capture pertinent post details and draw reasoned inferences about the poster's psychological state. This integration enables the generation of scientifically grounded and interpretable conclusions.

\noindent \textbf{Ablation Study 2: Impact of Discriminator.}
We created a model variant without the discriminator, which occasionally led to inaccurate or excessive explanations, resulting in erroneous judgments. While with the discriminator, the model initially misinterpreted the post. However, The discriminator acts as a form of third-party supervision, reminding and guiding the system to make accurate inferences. The performance comparison in Table \ref{Table: Abalation results} revealed the significant impact on improving the accuracy of mental health detection by incorporating the discriminator.

\section{Related Work}
Social media platforms have become a valuable tool for analyzing individual and collective psychological well-being and health trends. One important task is categorizing social media posts into various mental disorder classes, such as depression \cite{r3}. To achieve this, researchers have refined pre-trained language models (PLMs) such as BERT \cite{r4}, RoBERTa \cite{r5} and MentalBERT \cite{r6}. 

In recent years, there has been growing interest in the application of large language models (LLMs) in the field of mental health care \cite{r11, r12}. With the ability to retrieve and summarize information from extensive datasets, LLMs can interpret and predict behavioral patterns by analyzing textual data to identify indicators of mental health conditions or changes in psychological states. For example, fine-tuning LLMs on a dataset comprising mental health related social media posts can achieve accurate classification of depressive states, anxiety levels, and other mental health indicators \cite{r10} as well as develop a robust mental health prediction system \cite{r9}.

\section{Conclusion}
In this work, we have innovatively applied Large Language Models (LLMs) in conjunction with self-rating mental scales to identify mental health issues, automating a process traditionally performed manually. Our exploration into LLMs' ability to extract personality traits from non-descriptive text demonstrates their potential to replicate human-like responses in psychological assessments. By sidestepping the need for extensive labeled data, our approach introduces a more scientific method to mental health analysis, mirroring real-life psychological evaluation practices. The comparative analysis indicates that our model not only competes with but also, in some cases, exceeds SOTA performance across various methodologies, including some supervised methods. 
These results underscore the versatility and interpretability of LLMs, suggesting a significant advancement in mental health diagnostics that is both scientifically grounded and generalizable. 

\section{Ethical Considerations}
The dataset we use were collected from public media platforms, with strict adherence to privacy protocols and ethical principles to protect user’s personal information, our data sources exclusively rely on publicly available information.

\bibliographystyle{ACM-Reference-Format}
\balance
\bibliography{reference}
\end{document}